\documentclass{article}

     \usepackage[nonatbib,preprint]{neurips_2019}

\usepackage[utf8]{inputenc} % allow utf-8 input
\usepackage[T1]{fontenc}    % use 8-bit T1 fonts
\usepackage{hyperref}       % hyperlinks
\usepackage{url}            % simple URL typesetting
\usepackage{booktabs}       % professional-quality tables
\usepackage{amsfonts}       % blackboard math symbols
\usepackage{nicefrac}       % compact symbols for 1/2, etc.
\usepackage{microtype}      % microtypography
\usepackage{graphicx}
\usepackage{subcaption}

\title{ProdSumNet: reducing model parameters in deep neural networks via product-of-sums matrix decompositions}

\author{
  Chai Wah Wu\\
  IBM Research AI\\
  IBM T. J. Watson Research Center\\
  P. O. Box 218\\
  Yorktown Heights, NY 10598 \\
  \texttt{cwwu@us.ibm.com} \\
  May 23, 2019\\
}

\begin{document}

\maketitle

\bibliographystyle{abbrv}

\begin{abstract}
We  consider a general framework for reducing the number of trainable model parameters in deep learning networks by decomposing linear operators as a product of sums of simpler linear operators. Recently proposed deep learning architectures such as CNN, KFC, Dilated CNN, etc. are special cases of this framework and we illustrate other types of neural network architectures within this framework. We show that good accuracy on MNIST and Fashion MNIST can be obtained using a relatively small number of trainable parameters. In addition, since implementation of the convolutional layer is resource-heavy, we consider an approach in the transform domain that obviates the need for convolutional layers.  One of the advantages of this general framework over prior approaches is that the number of trainable parameters is not fixed and can be varied arbitrarily. In particular, we illustrate the tradeoff of varying the number of trainable variables and the corresponding error rate. As an example, by using this decomposition on a reference CNN architecture for MNIST with over $3\cdot 10^6$ trainable parameters,  we are able to obtain an accuracy of $98.44\%$ using only $3554$ trainable parameters. 
\end{abstract}

\section{Introduction}
In recent years, deep neural networks \cite{LeCun2015} have re-emerged as a powerful tool in many nonlinear regression and classification applications, especially in the visual \cite{Voulodimos2018} and audio \cite{Gavat2015} processing domains. The general form of a neural network can be expressed as a sequence of linear and nonlinear mappings. In particular, we consider the following $N$-layer feed-forward network formulation:

\begin{equation} y_{i+1} = f_i(W_iy_i + b_i), i = 1,..., N
\label{eqn:nn}
\end{equation}
where $f_i$ is a nonlinear function mapping vectors of length $m_i$ to vectors of lengths $n_{i+1}$. They can be vectorized activation functions (ReLu, sigmoid, tanh, etc.), but can also be other functions such as pooling, softmax and even linear functions such as the identity function. Each $W_i$ is a matrix of length $m_i\times n_i$ and $y_i$ is a vector of length $n_i$, although for easier visualization and interpretation, they can be represented as (or reshaped into) a multidimensional tensor with the same number of elements. The vectors $b_i$ are of length $m_i$.
The vectors $y_1$ and $y_N$ are the input and the output vectors of the entire network respectively and thus the network in Eq. (\ref{eqn:nn}) describes a nonlinear mapping between input and output. The functions $f_i$ are assumed to be fixed and given. The goal in training a network for classification and regression is that given $K$ pairs of vectors $(x_k, z_k)$, find the set of weights $W_i$ and biases $b_i$ such that $\sum_k d(\tilde{z}_k, z_k)$ is minimized, where $\tilde{z}_k$ is the output from the neural network when $x_k$ is fed as input to the neural network and $d(\cdot,\cdot)$ is an error or loss metric. 

\section{Model reduction}
There are 2 ways in which model reduction is beneficial in deep learning. In the first case, after the model weights are obtained via training, model reduction can be used to reduce the resources needed for inference. In this case, the fixed weight matrices are implemented via approximations such as SVD \cite{Xue2013}. In the second case, which this paper is focused on, we want to reduce the complexity of the deep network that is used during training. 
One main factor in determining the size and complexity of the network in Eq. \ref{eqn:nn} is the number of trainable weights $n_W$. For the commonly implemented case where every weight can be set independently, this is equal to as the sum of  the number of elements in all the matrices $W_i$ and vectors $b_i$.  When $n_W$ is large, this can slow down training by requiring more iterations and more computations per iteration. The goal of this paper is to present a general framework for reducing $n_W$ that encompasses many model reduction techniques. 
The most well known model reduction technique is the introduction of the convolutional layer in convolutional neural networks. By making the weights shift-invariant across the image, the number of independently trainable weights is drastically reduced. This technique works well in problem domains that embody this shift-invariance, and this include approaches that works well in the frequency transform domain, whos transforms are circulant (or block circulant with circulant blocks in the case where the data occurs naturally in 2D). Other recent techniques include replacing the dense weight matrices with circulant matrices \cite{Ding2017}.

\section{Model reduction via product-of-sums decompositions}
The main idea in this paper  is to reduce the full matrix $W_i$ and vectors $b_i$ with a parametrized version that has fewer parameters than the number of entries.
In particular, we consider the following decomposition of a matrix\footnote{Even though we focused on decompositions of matrices, this decomposition is also applicable to higher order tensors. In higher order tensors, the sum is well-defined as addition elementwise (as long as they have the same shape) and the product operation can be defined in various ways, e.g. as a tensor product, a $k$-mode tensor product \cite{Kolda2009}, as a t-product 
\cite{Kilmer2011} or a generalized t-product \cite{Kernfeld2015}.}
 as a product of sums:

\begin{equation}\label{eqn:decomp} W = \prod_{j=1}^p\sum_{k=1}^{s_i} g_{jk}(a_{jk})M_{jk}
\end{equation}
where $a_{jk}, j = 1,\cdots, p, k = 1,\cdots, s_j$, are the the {\em trainable parameters}, also denoted as the {\em variable parameters}. The total number of trainable parameters $n_W$ in Eq. \ref{eqn:decomp} is $\sum_j s_j $. The {\em fixed parameters} are the matrices $M_{jk}$ and the nonlinear functions $g_{jk}$. The size of the matrices $M_{jk}$ are such that the addition and product operations are valid, i.e.
$M_{jk}$ and $M_{jk'}$ have the same dimension and the product of $M_{jk}$ and $M_{j+1,k}$ is a valid operation\footnote{We use the convention that $\prod_i M_i = M_1M_2M_3\cdots$, i.e. increasing indices are to the right.}.  If the optimal weight matrix $W_{opt}$ has an (approximate) decomposition with a small number of parameters $a_{ij}$ (i.e. $\sum_i s_i$ is small), then this would reduce the model significantly. If a good approximation to $W_{opt}$ is known (either via transfer learning or training a small number of initial epochs), then a low rank decomposition can be computed to reduce the number of parameters. This approach was used in \cite{Sainath2013} for speech processing. Decompositions satisfying specific sparsity patterns of $M_i$ are explored in \cite{Wu2011}.  We will restrict ourselves to the case $g_{jk}(x) = x$ in the sequel since the case of nonlinear $g_{jk}$ can be easily addressed when $g_{jk}$ are differentiable.

It is clear that convolutional layers \cite{chua-yang88,LeCun2015}, KFC \cite{Grosse2016}, circulant matrices, Toeplitz matrices, Handel matrices, etc. \cite{Zhao2017}, all can be decomposed as Eq. (\ref{eqn:decomp}). The difficult part is to find the fixed parameters and the decompositions that would work best while keeping the $n_W$ small. The proposed framework offers more flexibility in choosing these parameters and the number of them. For instance, to approximate an $n\times n$ matrices using circulant matrices or the Adaptive Fastfood Transform \cite{Yang2015}, the number of parameters are fixed at $n$ and $3n$ respectively, whereas in the current approach, the number of trainable parameters can be varied arbitrarily, as we will illustrate in subsequent examples.

Note that this decomposition can be easily implemented in current deep learning architectures. The weight matrix is a linear with respect to the trainable parameters, and thus its partial derivative with respect to the error will be related by the same linear operator. For example, for $W$ decomposed as a sum of matrices $M_k$, and a layer is described as $ y = f(\sum_{k}a_k M_k x + b)$, then $\frac{\partial y}{\partial a_k} = \nabla f( \sum_{k}a_k M_k x + b)M_kx$. Product of matrices are equivalant to multiple layers in a neural network where the nonlinearity is the identity function and its partial derivative for $y = f((\prod_i\alpha_i M_i) x+b)$ is equal to
$\frac{\partial y}{\partial a_j} = \nabla f((\prod_i \alpha_i M_i) x + b) \prod_{i\neq j} \alpha_i (\prod_i M_i) x$.

There are several  benefits in reducing the number of trainable parameters. The first is that the gradient and Hessian (or approximate Hessian in quasi-Newton methods) are smaller objects and this can speed up training. The second is that the number of trainable prameters can be viewed as knobs adjusting the strengths of a small number of operators described by $M_{jk}$ and $b_i$. For a single fixed matrix $W_i$ that have been obtained after training, there exists hard coded approaches using physics \cite{Lineaat8084} or in silicon (using ROM or FPGA). If each of the $M_{jk}$ and $b_i$ are hard coded using these approaches then we can build a hardware-based classifier than can be trained using much fewer variable memory. Another benefit is that there is less chance of overfitting and thus reducing the need of a dropout layer. On the other hand, there is a tradeoff in performance as the model reduction reduces the degree of freedom.

\section{Decomposition as a sum of matrices: $p=1$}
In this section we consider the simplest case where the matrix $W$ is decomposed as a linear combination of matrices with the linear coefficients being the trainable parameters, i.e. the decomposition can be reduced to:

\begin{equation}
\label{eqn:lindecomp}
W = \sum_{i=1}^s a_i M_i
\end{equation}

What should $s$ be, i.e. how many trainable parameters are needed? If $M_1$ is close to the optimal weight matrix $W_{opt}$, then we can trivially choose $s=1$. In practice, the optimal matrix $W_{opt}$ is unknown and the goal is to find a small set of matrices $M_i$ such that the $W_{opt}$ for the problem at hand is likely in the linear subspace of these matrices.
Eq. (\ref{eqn:lindecomp}) can be rewritten as follows in order to be easily implemented in numpy and TensorFlow  using the \verb|tensordot| operator:

\begin{equation}
\label{eqn:lindecomp_matrix}
W = [M_1| M_2 | \cdots | M_u] (A\otimes I) 
\end{equation}
where $A$ is the column vector $[a_1, \cdots a_u]^T$.

In dimensionality reduction via random projection\cite{Dasgupta2000}, the input row vector $x$ of order $n$ is multiplied (on the right) by a matrix $W'$ of order $n\times m$ where $m <<  n$ in order to form a much smaller row vector of order $m$ which is then used as input to a neural network with a full dense layer with weight matrix $B$ of order $m\times k$ where $k$ is the number of features in the input layer. This is equivalent to decomposing the full matrix $W$ as $W'B$ where all entries of  $B$ are the trainable parameters.
This can be rewritten in the form of Eq. (\ref{eqn:lindecomp}) as $W  = W'B = \sum_{ij}b_{ij}M_{ij}$
where $b_{ij}$ is the $ij$-th entry of $B$, $E_{ij}$ is a $0-1$ matrix of order $m\times k$ whose $ij$-th element is 1 and $0$ elsewhere and $M_{ij} = W'E_{ij}$.
Next we will illustrate this decomposition with an example application.

\subsection{Example 1: linear regression on MNIST dataset}
We consider a single layer neural network to classify the MNIST handwritten digits dataset \cite{LeCun2010}. As is well known, the accuracy on the testing set is approximately 92\%.
The input are 28 by 28 grayscale images and the output is a one-shot encoding of the classified digit which is decoded using softmax. Thus the weight matrix $W$ is a 784 by 10 matrix and the bias vector $b$ is a 10 by 1 vector resulting in 7850 trainable parameters (a model which we denote as the reference model\footnote{In the subsequent examples, we will pick as the reference model a classifier architecture in the literature where the weight matrices $W_i$ are not decomposed as Eq. (\ref{eqn:decomp}) and each entry is a trainable parameter.}). We consider 3 ways to reduce the number of model parameters of $W$ as a sum of matrices. In all these cases, we construct a fixed matrix $W'$ of size $784\times m$. The trainable parameters are the entries of a matrix $B$ of order $m\times 10$ and a vector $b$ of length $10$ and $W = W'B$ resulting in a total of $10(m+1)$ trainable parameters.

In the first case, $W'$ consists of the map from the image to the $m$ lowest 2-d DCT (Discrete Cosine Transform) coefficients. This is a reasonable choice since image compression algorithms such as JPEG \cite{pennebaker:jpeg1993} leverage the insight that the lowest DCT coefficients contain most of the information relevant to the human observer.   In the second case, $W'$ is random matrix with entries taken independently from a standard Gaussian normal distribution as is typically used in random projection methods to reduce dimensionality \cite{Dasgupta2000}\footnote{In the sequel, all random matrices used are generated this way.}. In the third case, $W'$ consists of $m$ columns of a random permutation matrix. This corresponds to a random subsampling of $m$ pixels of the input image to be used for classification.

\begin{figure}[htbp]
\centering
\begin{subfigure}[b]{0.40\textwidth}
\centerline{\includegraphics[width=\textwidth]{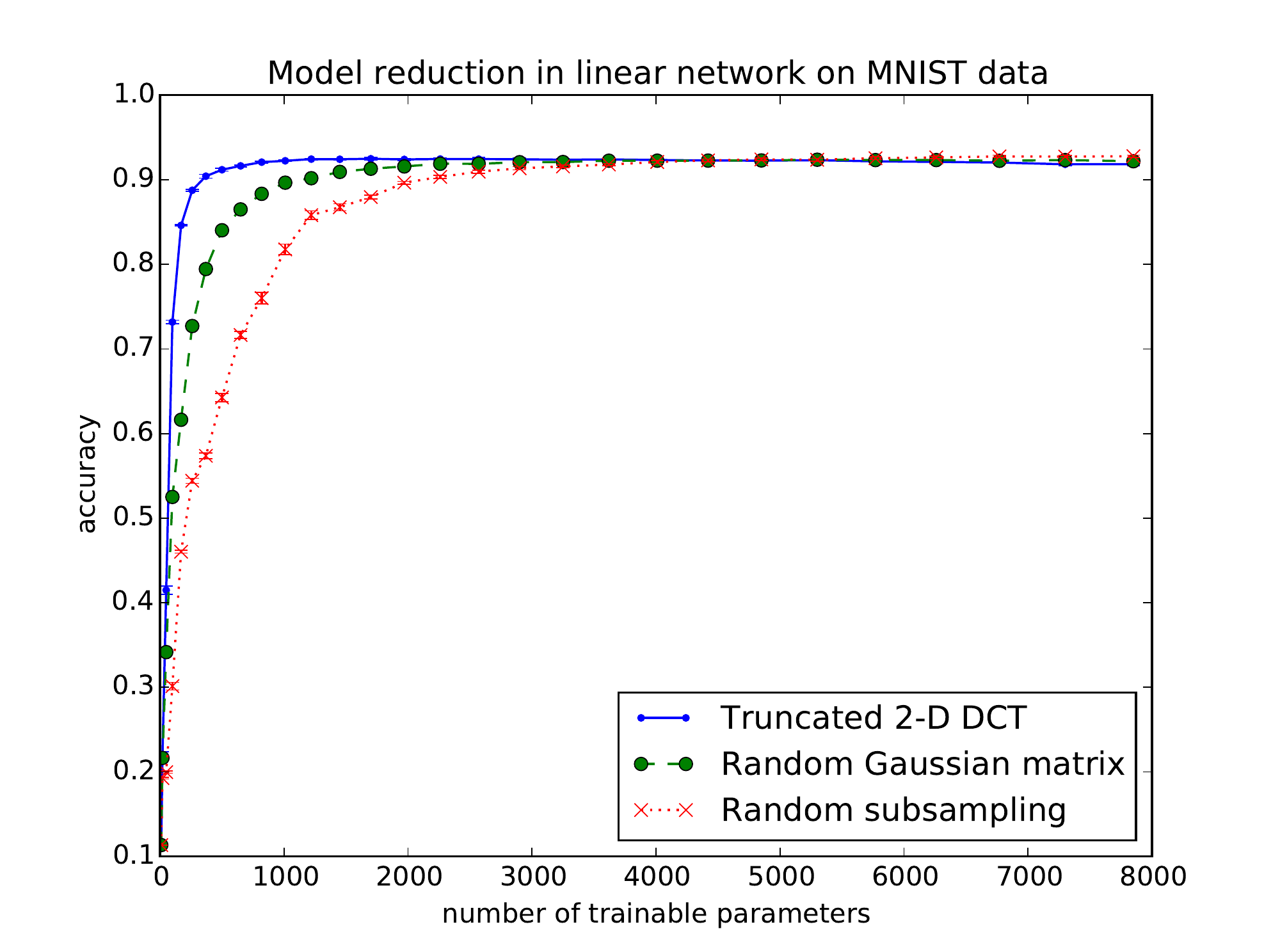}}
\caption{MNIST data.}\label{fig:MNIST-linear}
\end{subfigure}
\quad
\begin{subfigure}[b]{0.40\textwidth}
\centerline{\includegraphics[width=\textwidth]{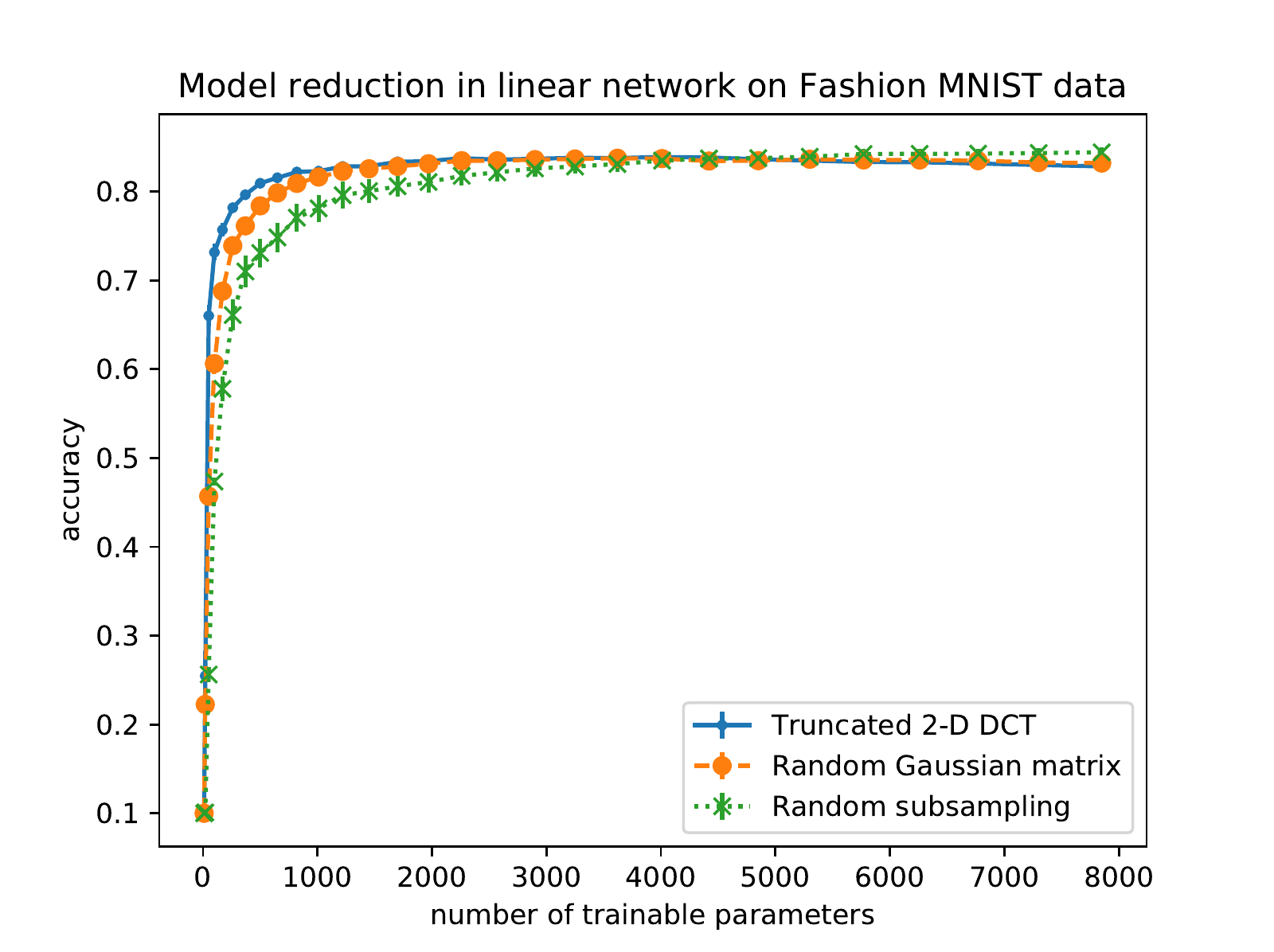}}
\caption{Fashion MNIST data.}\label{fig:fashionMNIST-linear}
\end{subfigure}
\caption{Model reduction of the linear model, averaged over 10 trials. The reference model has 7850 trainable parameters. }
\end{figure}

The results are shown in Fig. \ref{fig:MNIST-linear}. We see that with about 10\% of the total number of model parameters of the reference model, we can achieve similar performance when the truncated 2-d DCT matrix $W'$ is used. In fact, using about 22\% of coefficients we obtain a slightly better error rate on the testing data than when all the coefficients are used, indicating either overfitting due to the higher frequency coefficients containing minutiae and fine details or slower convergence to the optimal weights when the full dense matrix $W$ is trained. The performance for the random Gaussian matrix is lower, but again a subset of parameters (about a third) is sufficient to achieve similar performance as the full matrix. Finally, the performance of the reduced model is the worst for random subsampling requiring samples from more than half the image before similar classification performance is achieved.

We also perform the same experiment on Fashion MNIST data \cite{xiao_fashion-mnist:_2017} with similar behavior (Fig. \ref{fig:fashionMNIST-linear}). This illustrates that for both the MNIST and Fashion MNIST classification tasks, about 10\% of the DCT  coefficients of the image is sufficient to achieve the linear classifier performance.

\section{Decomposition into products of multiple sums: $p > 1$}
In the prior example we replace a dense matrix where each element is a trainable parameter with a linear combination of $m$ fixed matricess where the linear coefficients are the trainable parameters. 
In thise section, we look at the case of multiplying several such sums to form the dense matrix.

\subsection{Linear classifier for MNIST}
We perform an experiment using $10$ products, i.e. $p = 10$ with $s_i = K$ for all $i$ and picking the entries of matrices $M_{jk}$ from a normal distribution. The number of trainable parameters is $10(K+1)$. If $W$ has order $n\times m$, we choose $M_{jk}$ to be a square matrix of order $n\times n$ for $j=1,\cdots, p-1$ and
$M_{jk}$ to be the same order as $W$ for $j=p$. The results are shown in Fig. \ref{fig:MNIST-prodsum}, which includes the data from Fig. \ref{fig:MNIST-linear}. We see that the performance is superior to a single product $(p=1)$ and approaches that of the DCT. Notice that due to the Fast Cosine Transform, the DCT can be decomposed into a product of sparse matrices. It is remarkable that product of random matrices can approach the performance of the truncated DCT which leverages the inherent knowledge of spectral decomposition and the observation that visual tasks can be performed using only a compressed version of the image (corresponding to using only the low frequency DCT components). We expect this product of sums architecture to perform well in non visual (or auditory) classification tasks.

\subsection{Example 2: MNIST and Fashion MNIST using LeNet}

We consider the LeNet \cite{LeCun1988} architecture, but using ReLu for the activation function instead and using the parameters as described in the Tensorflow tutorial "Deep MNIST for Experts". It consists of 2 convolutional layers, each followed by a maxpooling layer, and followed by a fully connected dense layer, a dropout layer and another fully connected dense layer.
We obtain with this reference model an accuracy on the testing dataset of approximately $99.34\%$ and $92.5\%$ for MNIST and Fashion MNIST respectively. The total number of trainable parameters in this reference model is 3274634, with most of them in the 2 fully connected dense layers. In this section we look at various attempts of reducing the number of trainable parameters, while trading off test accuracy.

\subsubsection{Convolutional feature extraction layers}
\begin{figure}[htbp]
\centering
\begin{subfigure}[b]{0.45\textwidth}
\centerline{\includegraphics[width=\textwidth]{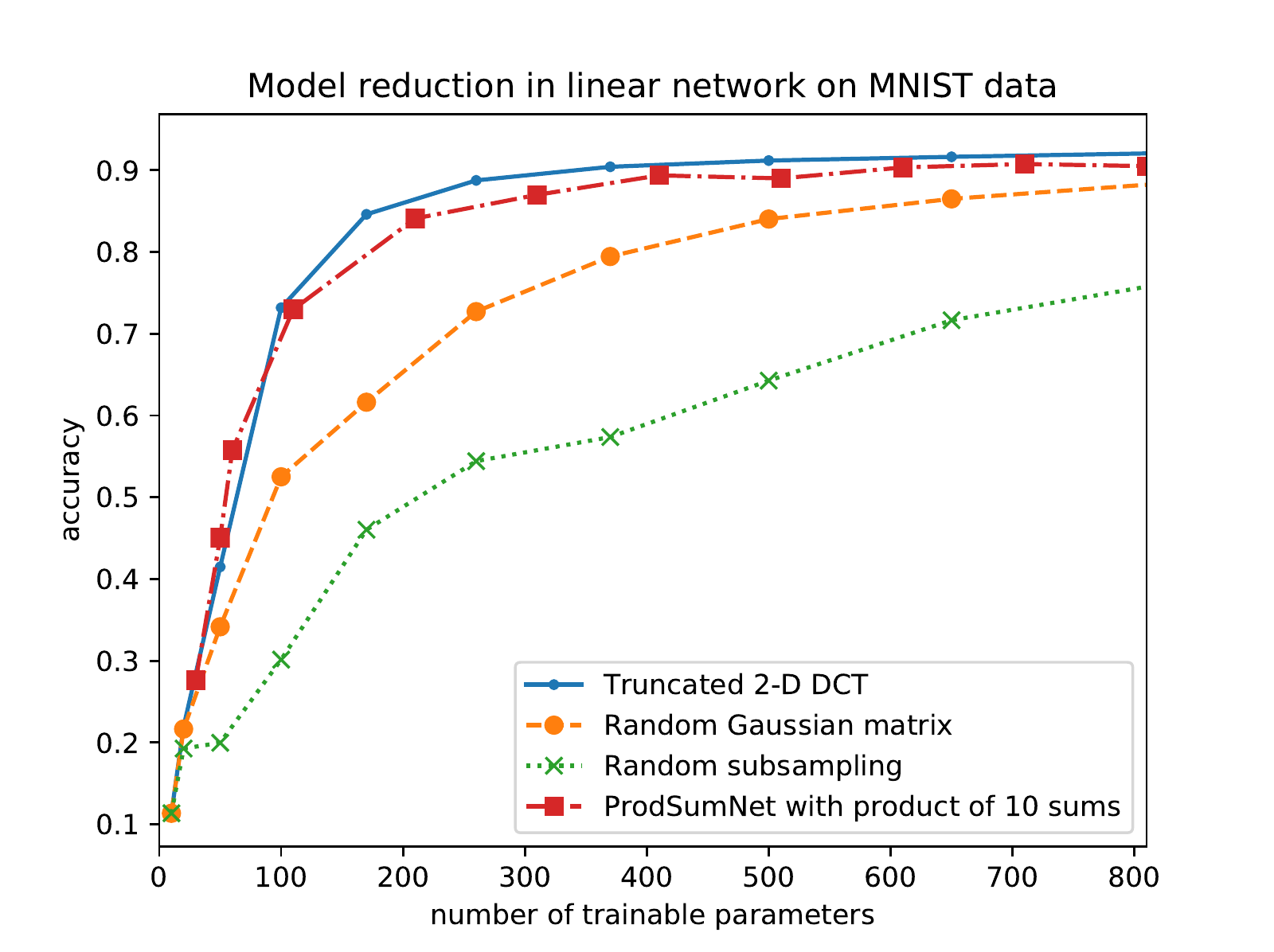}}
\caption{Model reduction in a linear model classifying MNIST data using a product of 10 sums of $K$ matrices for various $K$. The reference model  has a total of 7850 trainable parameters.}\label{fig:MNIST-prodsum}
\end{subfigure}
\quad
\begin{subfigure}[b]{0.45\textwidth}
\centerline{\includegraphics[width=\textwidth]{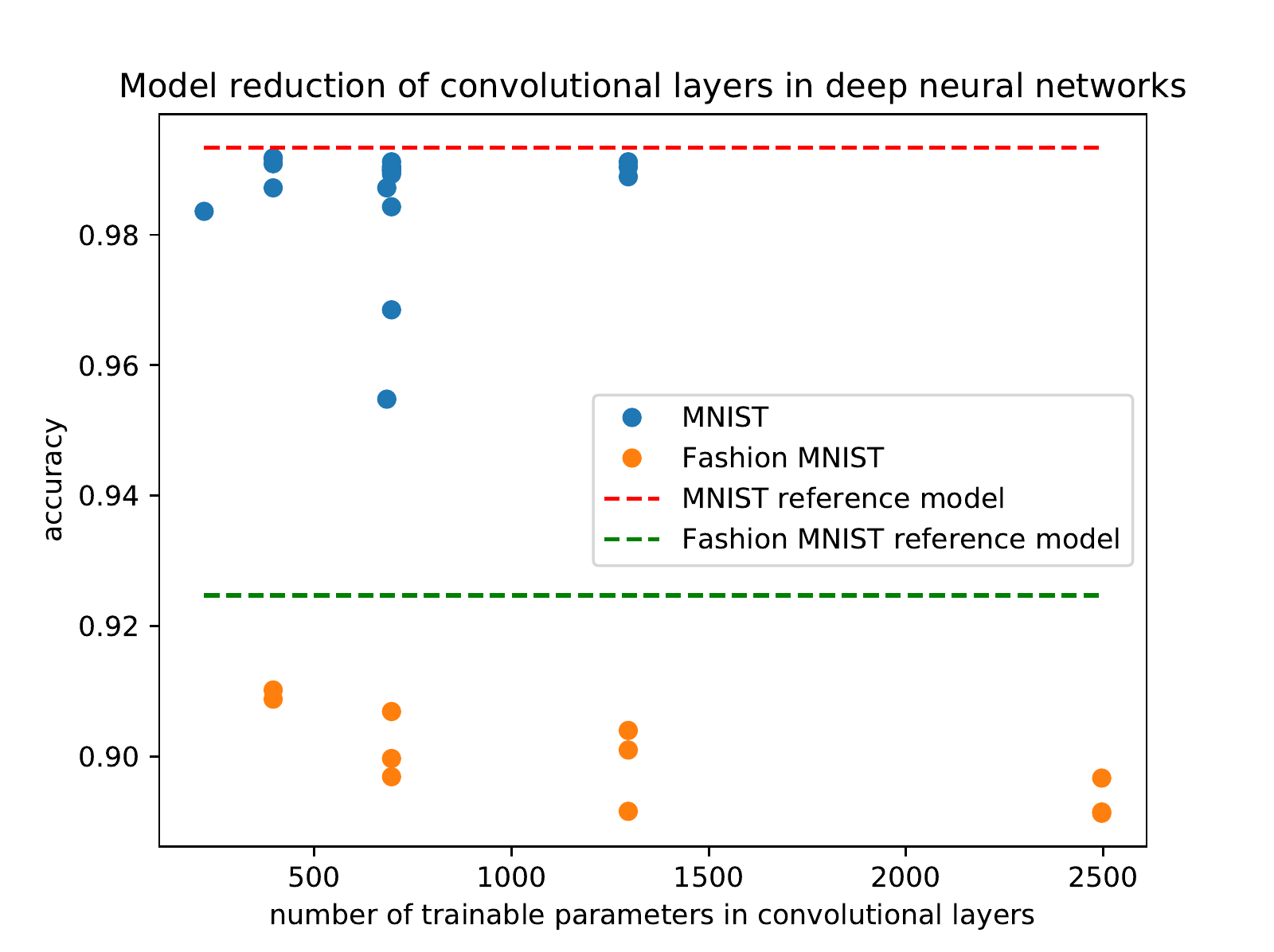}}
\caption{Model reduction of the 2 convolutional layers in classifying MNIST and Fashion MNIST data. The reference model's convolutional layers has a total of 52096 trainable parameters.}\label{fig:MNIST-cnn}
\end{subfigure}
\caption{}
\end{figure}

The first convolutional layer is a $5\times5$ filter generating $32$ features corresponding to $(5\times 5 + 1)\times 32 =  832$ parameters, whereas the second convolutional layer is a $5\times 5$ filters taking $32$ features and mapping them to $64$ features resulting in $(5\times 5\times 32+1)\times 64 = 51264$ parameters, thus the $2$ convolutional layers has a total of $52096$ trainable parameters.
We replace each convolutional layer with a product of weighted sums of random matrices for various values of $s_j$ and $p$ and matrices $M_{jk}$. As shown in Fig. \ref{fig:MNIST-cnn}, similar performance as the reference model can be obtained with far fewer trainable parameters in the convolutional layers on the MNIST dataset.   On the Fashion MNIST dataset, an accuracy to within about 1\% with less than 1\% of the trainable parameters in the convolutional layers.

Since the convolutional layers encompass only a small fraction of all trainable parameters, reducing the number of parameters in the convolutional layers may not seem like a useful approach in simplifying the network. However, this decomposition effectively decomposed the convolution into a weighted sum of a relatively small number of fixed {\em basic} or {\em core} filters and each such filter can be hardcoded into an optical filter and the processing can be performed in parallel and at the speed of light \cite{Weaver1966}.  This is important as the implementation of the convolutional layers in a trained classifier takes up most of the time on a traditional computer.

\subsubsection{Dense analysis layers}
\begin{figure}[htbp]
\centering
\begin{subfigure}[b]{0.45\textwidth}
\centerline{\includegraphics[width=\textwidth]{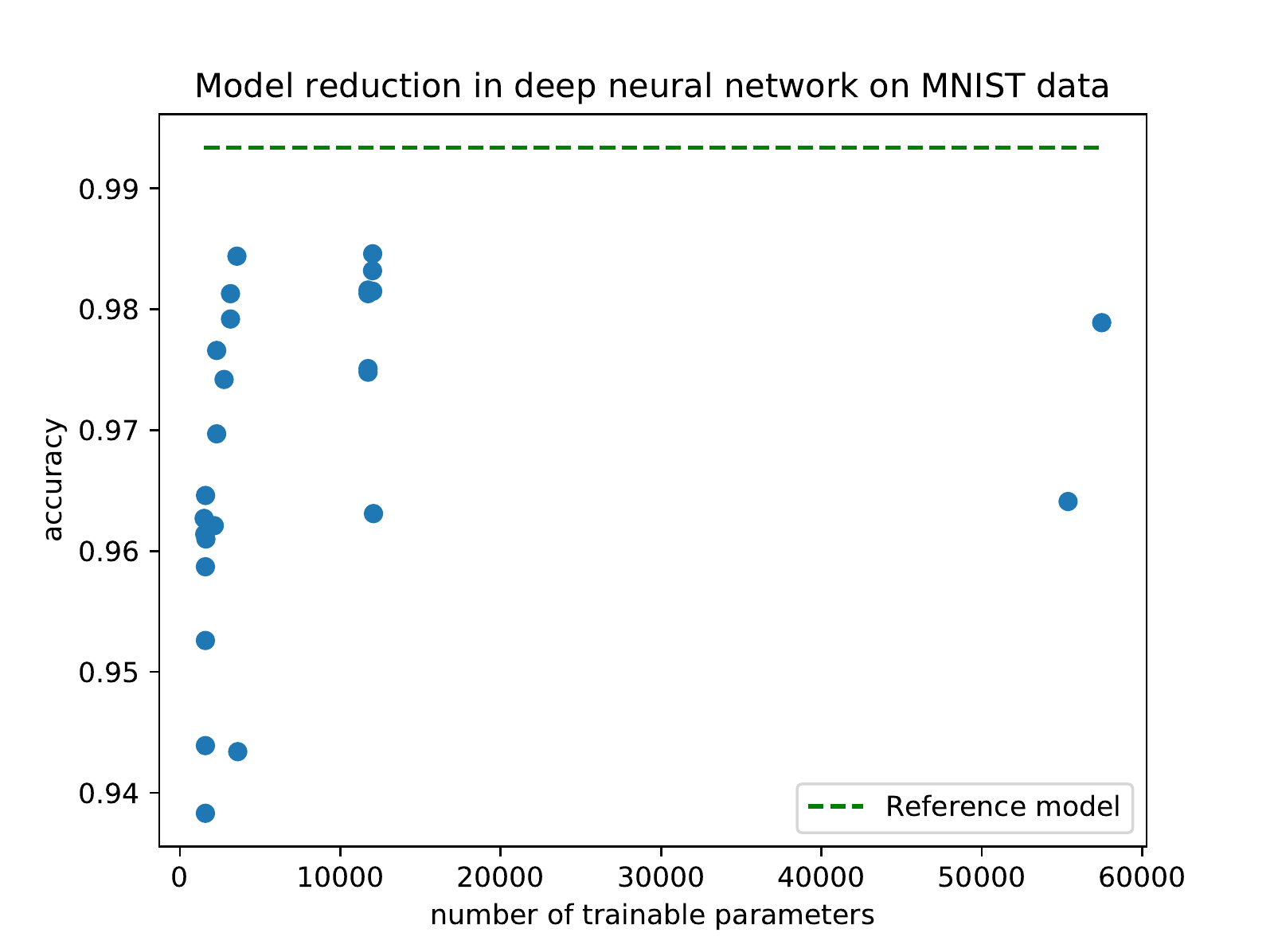}}
\caption{MNIST data.}\label{fig:MNIST-dense}
\end{subfigure}
\begin{subfigure}[b]{0.45\textwidth}
\centerline{\includegraphics[width=\textwidth]{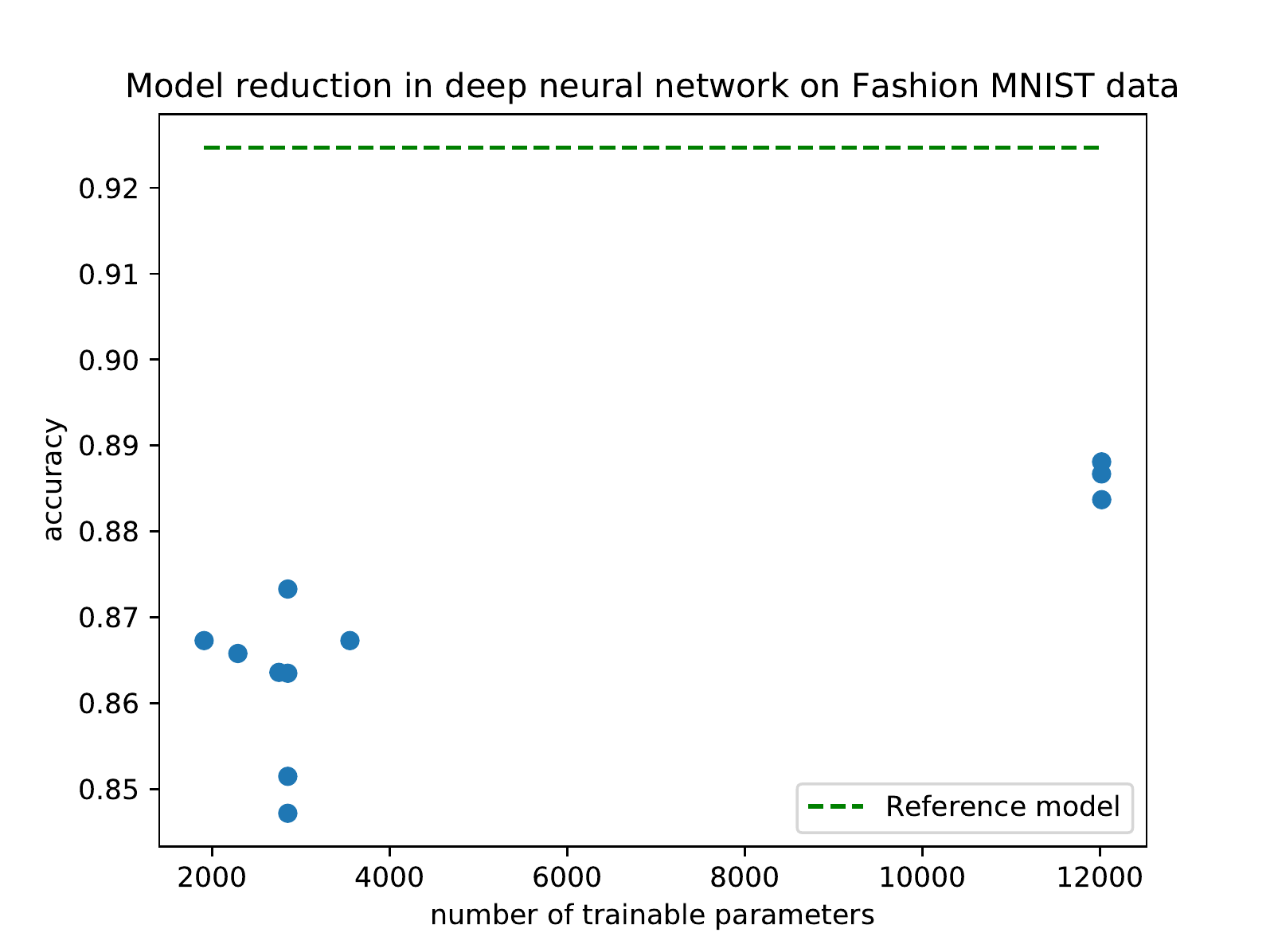}}
\caption{Fashion MNIST data.}\label{fig:FMNIST-dense}
\end{subfigure}
\caption{Model reduction replacing both convolutional layers and full connected layers with products of sums. The reference model  has a total of over $3\cdot 10^6$ trainable parameters.}
\end{figure}

By replacing both the convolutional layers and the fully connected layers (that have the most trainable parameters) with various ways to decompose the weight matrix as Eq. (\ref{eqn:decomp}). In particular, we consider various combinations of $p$ and $s_i$ and also pick $M_{ik}$ to be both dense random matrices and rank-1 matrix of the form $u\otimes v$ where $u$ and $v$ are random vectors. The results are shown in Fig. \ref{fig:MNIST-dense}. We see that the accuracy for classifying MNIST has decreased, but with far fewer trainable parameters. We see that we get to within about $1\%$ of the accuracy of the reference LeNet architecture using only $0.1\%$ of the number of trainable parameters. In particular, using $200$ and $800$ fixed random matrices for the two convolutional layers, $100$ full rank plus $500$ rank-1 random matrices for the first dense layer and $1024$ random matrices for the second dense layers, we obtained a network with $3554$ trainable parameters and a test accuracy of $98.44\%$. The results for Fashion MNIST  (Fig. \ref{fig:FMNIST-dense}) is a little worse, but still close to within $4\%$ with only $0.38\%$ of the number of parameters. 

Table \ref{tbl:comparison} compares this approach and the tradeoff in accuracy with Adaptive Fastfood Transform and KFC on MNIST. Since these prior approaches are special cases of Eq. (\ref{eqn:decomp}), and the number of trainable parameters is constrained by the special structure of the decomposition, the current framework provides a more flexible tradeoff between the number of trainable parameters and accuracy.

\begin{table}
\begin{center}
\begin{tabular}{|l|r|r|}
\hline
{\bf Deep learning network architecture} & {\bf \# of trainable params} & {\bf Error} \\
\hline\hline
Adaptive Fastfood 2048  \cite{Yang2015} & 52124 & 0.73\%\\
\hline
Adaptive Fastfood 1024 \cite{Yang2015} & 38821 & 0.72\%\\
\hline
KFC-Combined \cite{Grosse2016} & 52500 & 0.57\% \\
\hline
KFC-II \cite{Grosse2016} & 27700 & 0.76\% \\
\hline
ProfSumNet (Fig. \ref{fig:MNIST-dense}) & 3554 & 1.55\%\\
\hline
\end{tabular}

\end{center}
\caption{Comparison with other methods of reducing trainable parameters on MNIST. Both Adaptive Fastfood Transform and KFC can be considered special cases of Eq. \ref{eqn:decomp} with $M_{jk}$ taking specific forms.}
 \label{tbl:comparison}

\end{table}

\subsection{Multi Layer Perceptron (MLP)}
We consider a 3-layer neural network with dense connected layers as a reference model for MLP. The 3 layers have 1024, 512 and 10 neurons respectively with a ReLu activation function layer in between two layers. This network achieves 98.2\% accuracy on MNIST with 1333770 trainable parameters. We reduce the network by combining the decompositions described earlier. In particular,  the first layer is reduced using the low frequency DCT coefficients, the second layer using a products of sums of random matrices and the third layer is either kept as a dense layer or decomposed as a product of sums. The performance is shown in Fig. \ref{fig:MNIST-mlp}. We see that we can achieve within 1.5\% of the performance of the reference model using only 4.3\% of the number of trainable parameters.

\section{Deep learning for vision applications without convolutional layers}
Even though convolutional layers appears crucial in modern visual recognition deep learning architectures, there are some issues with such architectures.
First of all, the execution of the convolutional layer during classification is time consuming. Since convolution is equivalent to Hadamard product in transform space, this is exploited in \cite{Mathieu2014} to speed up the computation by transform the data to fourier space, apply the filtering via Hadamard product and transform the output back via the inverse fourier transform.  
Secondly, the filter size is fixed and relatively small, as the number of trainable parameters is proportional to the filter size.
In this section we consider an approach of operating the nonlinearity entirely within the frequency transform domain.  This approach addresses both the issues above. First of all, it allows the use of weight matrices, whose corresponding filter in the spatial domain is not restricted to a specific filter size. Secondly, the use of the Hadamard product speeds up the computation considerably, especially when the desired filter size is relatively large. Each filter can be considered has a Hadamard product with a diagonal matrix.
We construct a neural network consisting of the following layers:

\begin{enumerate}
\item DCT layer: each image is converted to the 2D DCT via a matrix multiply
\item Hadamard product layer 1 with 8 filters corresponding to 8 Hadamard products. This can be viewed as 8 diagonal matrix. We construct them as sum of 100 randomly generated 8-tuples of diagonal matrices. A bias vector is also added
\item  ReLu nonlinearity layer
\item Reduction layer: a linear layer to map vectors of size $w\times h$ to vectors of size $w\times h/4$.
\item Hadamard product layer 2 with 8 filters constructed  as sum of  200 randomly generated 8-tuples of diagonal matrices. A bias vector is also added
\item  ReLu nonlinearity layer
\item  Dense layer 1: maps to 2048 features
\item ReLu nonlinearity layer
\item Dense layer 2: maps to 10 features
\end{enumerate}

\begin{figure}[htbp]
\centering
\begin{subfigure}[t]{0.45\textwidth}
\centerline{\includegraphics[width=\textwidth]{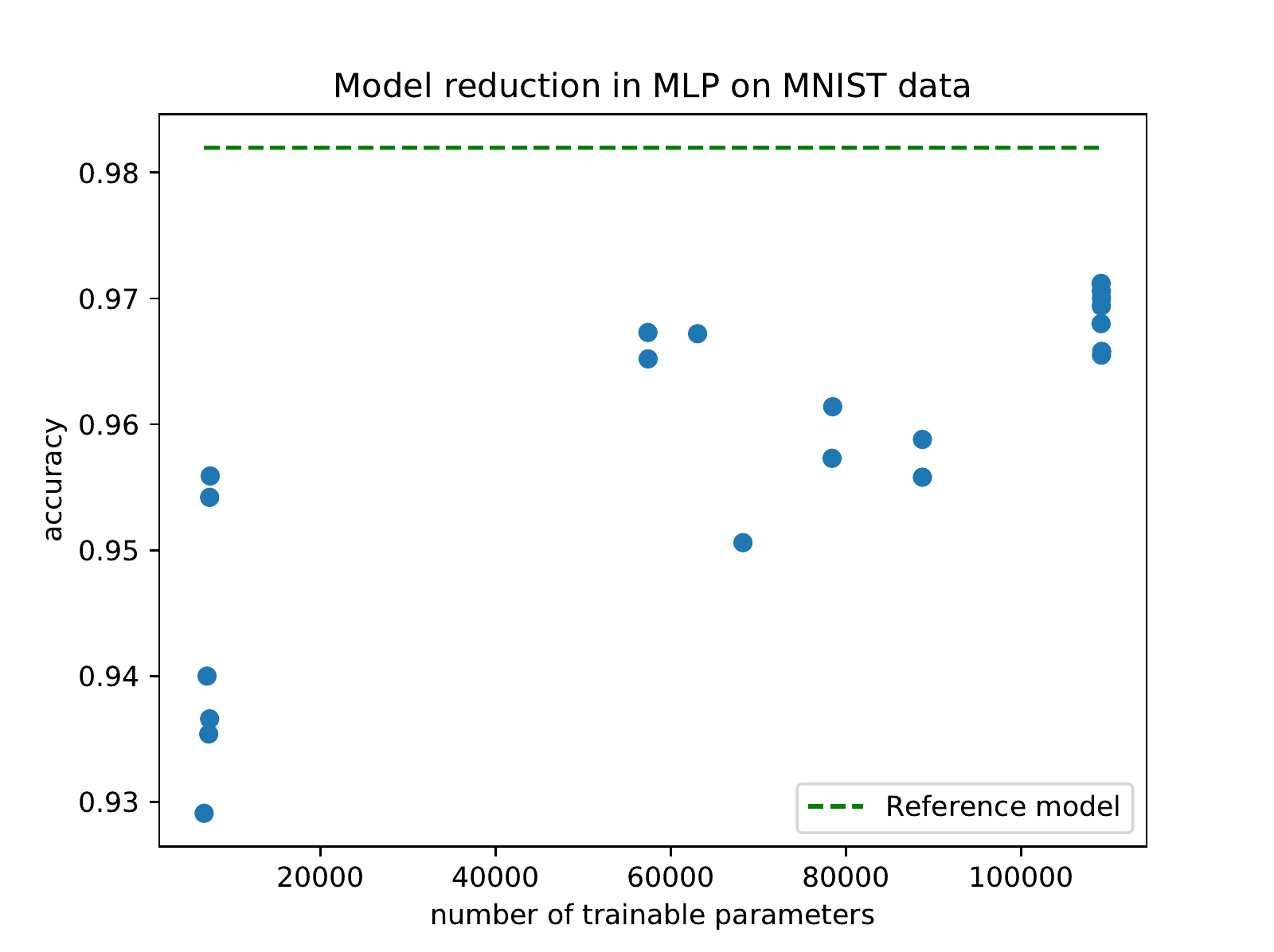}}
\caption{Model reduction in classifying MNIST data using a 3-layer MLP. The reference model  has a total of over $1.3 \cdot 10^6$ trainable parameters.}\label{fig:MNIST-mlp}
\end{subfigure}
\quad
\begin{subfigure}[t]{0.45\textwidth}
\centerline{\includegraphics[width=\textwidth]{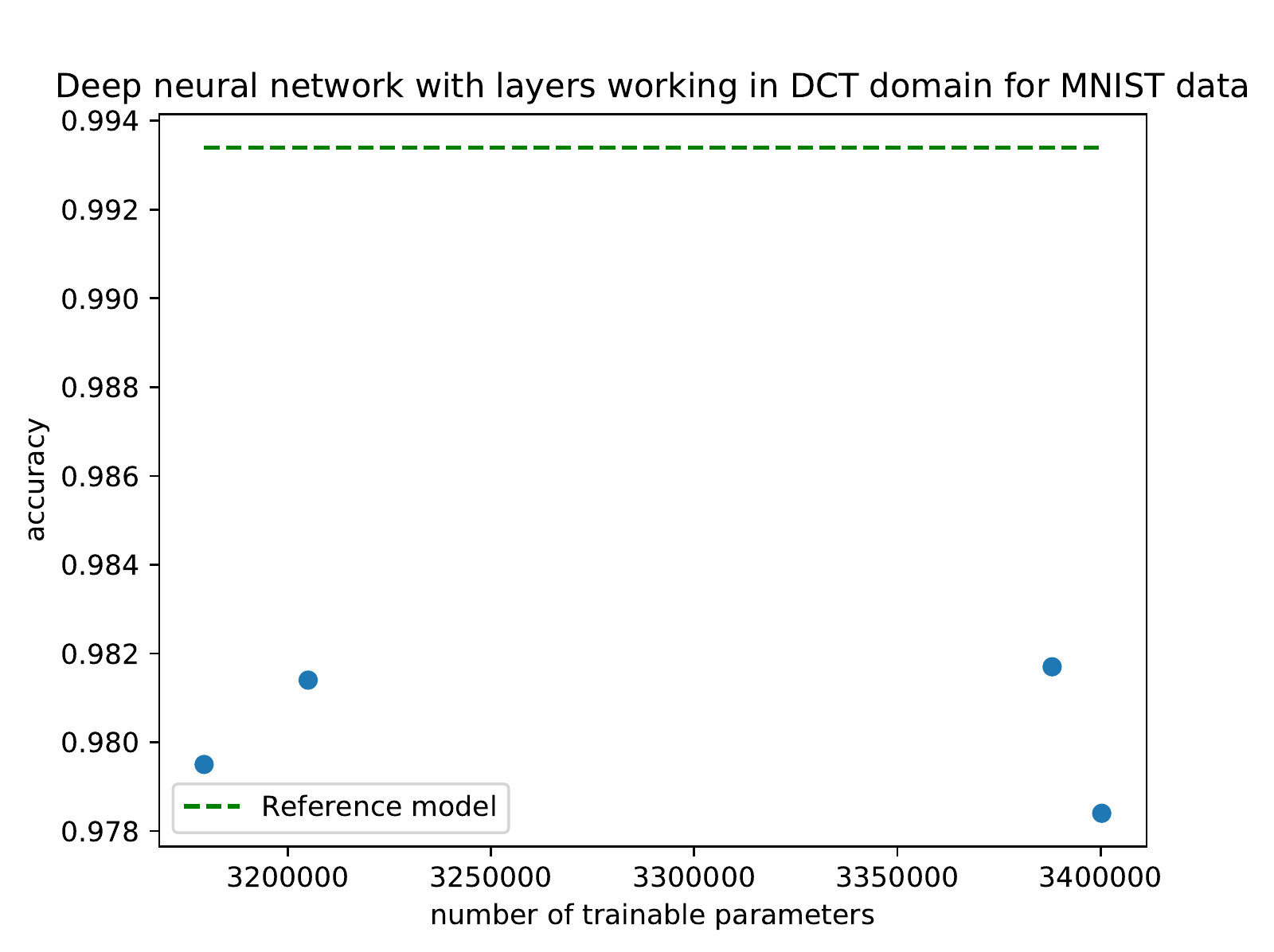}}
\caption{Deep neural networks using filters in the DCT domain classifying MNIST data.}\label{fig:MNIST-noconv}
\end{subfigure}
\caption{}
\end{figure}

Using these parameters, the Hadamard layers contains only $300$ trainable parameters, whereas the dense layers are similar (in terms of trainable parameters) to the reference network.
We see in Fig. \ref{fig:MNIST-noconv} that we can get $98.2\%$ accuracy without the use of a convolutional layer and thus can be implemented much more quickly.
A similar approach is discussed in \cite{Pratt2017}, but the FFT is used instead of DCT and the convolutional layers are matrices whose size is equal to length of the images $\times $ the number of features, thus fixing the number of trainable parameters to be relatively large and limiting the size of the images that can be processed. For instance, the approach in \cite{Pratt2017} reaches an error rate of $2.5\%$ for MNIST, whereas we obtain an error rate of $1.8\%$.

\section{Training weights in stages}
The same weight matrix $W$ can have different decomposition with different number of trainable parameters. This suggests a training framework in stages where the same weight matrix is trained using successive decomposition with increasing number of trainable parameters.  
For instance, in Eq. (\ref{eqn:decomp}) suppose that in different stages, $p$ is constant, but the $s_i$'s are increasing in each stage. After training using the decomposition in one stage, the sums in the product can be approximated in the next stage efficiently by means of least squares. results in $W'$.  This set of parameters can be used as initial conditions for training using the next stage's  decomposition, etc.

\section{Storage and implementation issues}
When the matrix $W$ and $n_W = \sum_j s_j$ is relatively large, storing many large fixed matrices might overwhelm the storage requirements of the network. In the experiments above, this is addressed by picking $M_{jk}$ to be a small set of full rank matrices, in addition to lower rank matrices which can be stored more compactly, e.g. as sum of outer products. 
One appeal of using circulant matrices, Haar matrices, or low rank matrices for $M_i$ in existing literature is that they can be easily described. This typically
translate to requiring less storage. 
In \cite{ailon2009} the Fast John-Lindenstrass Transform is used to implementation a random matrix multiplication that is fast and require less storage.
An interesting question is how to choose a sequence of $M_i$ that are less "complex" and whose linear combination is rich enough to provide a good coverage. Perhaps a complexity-bounded Johnson-Lindenstrauss Theorem is needed.

How should we choose $M_i$ that are less complex? One possibility is to find $M_i$ that has low Kolmogorov complexity and generate them on the fly using the smallest program necessary, but this approach would increase the computation time.
Another approach is to have these matrices or their generators hard-coded in hardware which would result in faster execution as a side effect.

\section{Degrees of freedom and complexity of problem.}
Figs. (\ref{fig:MNIST-linear})-(\ref{fig:fashionMNIST-linear}) suggest that most of the information needed for classification of MNIST resp. Fashion MNIST resides in a small number of features. This is consistent with the image compression paradigm; images contain enough redundancies that can be removed without impacting its visual appearance and hence its recognition. The fact that in the JPEG compression standard, the high frequency DCT coefficients are less important, is consistent that a small number of DCT coefficients are sufficient to achieve high accuracy on MNIST and Fashion MNIST.
Given a specific problem domain, an interesting question is what is a small set of $M_i$ that can cover the range of weight matrices while trading off accuracy.
This analogy with compression goes further. Suppose we have multiple data sets besides MNIST and Fashion MNIST and train a certain CNN architecture (such as LeNet) on them, each resulting in a specific set of weight matrices. For simplicity, let us assume there is only one trained weight matrix $W_i$ in the network. The index $i$ is determined by the data set used to train it. Given a fixed $n_W$, what  would be the set of products of sums of such matrices that is closest to these weight matrices. More specifically, focusing on a specific weight matrix in the CNN architecture, and let $W_i$ be the trained weight matrix (which is not unique and depends on initial conditions) for dataset $S_i$, define $d(W_i,\{M_{jk}\})$ as
\[  d(W_i,\{M_{jk}\}) =  \min_{a_{jk}} \left\|W_i-\prod_{j=1}^p\sum_{k=1}^s g_{jk}(a_{jk})M_{jk}\right\| \]
where $\|\cdot\|$ is generally the Frobenius matrix norm.
Then we want to find $M_{jk}$ such that $\mu(\{d(W_i,\{M_{jk}\})\}_i)$ is minimized, where $\mu$ can be the mean or the maximum.
The experimental results shown in Figs. (\ref{fig:MNIST-cnn}-\ref{fig:FMNIST-dense}) indicate that the decrease in performance is larger for Fashion MNIST than MNIST suggesting that Fashion MNIST has more degree of freedom and requires more parameters to properly solve the classification problem.

\section{Concluding remarks} 
We show that decomposition of a matrix as a product of sums can be useful in reducing the complexity of the model and can lead to faster implementations. The proposed approach allow for a flexible tradeoff between the number of trainable parameters and the accuracy of the network. On the other hand, the choice of such a decomposition is crucial and an efficient way to search for the best decomposition is important. The experiments also indicate that the number of trainable parameters (i.e. the degrees of freedom) in traditional deep networks is more than necessary, leading to overfitting and the need for drop out layers. Finally, product of sums of random matrices appear to be a promising architecture that is applicable to general problem domains, as contrasted with CNNs which are suitable mainly for image processing tasks. Future work include developing efficient methods for finding the best decomposition.

\bibliography{cnn2,coding_theory,sensors,quant,markov,consensus,secure,synch,misc,stability,cml,algebraic_graph,graph_theory,control,optimization,adaptive,top_conjugacy,ckt_theory,math,number_theory,matrices,halftoning,image_processing,ai,communications,cad,sequences,computing,neural_nets}

\end{document}